\documentclass{sigchi}

\usepackage{balance}       
\usepackage[T1]{fontenc}   
\usepackage{txfonts}
\usepackage{mathptmx}
\usepackage{hyperref}
\usepackage{color}
\usepackage{booktabs}
\usepackage{textcomp}

\usepackage{subcaption}
\usepackage{graphicx}
\usepackage{lipsum}

\usepackage{float}
\restylefloat{table}

\usepackage{microtype}        
\usepackage{ccicons}          

\usepackage{todonotes}

\def\plaintitle{Efficient Non-uniform Quantizer for Quantized Neural Network Targeting Re-configurable Hardware}
\def\plainauthor{First Author, Second Author, Third Author,
  Fourth Author, Fifth Author, Sixth Author}

\def\plainkeywords{Authors' choice; of terms; separated; by
  semicolons; include commas, within terms only; required.}

\makeatletter
\def\url@leostyle{%
  \@ifundefined{selectfont}{
    \def\UrlFont{\sf}
  }{
    \def\UrlFont{\small\bf\ttfamily}
  }}
\makeatother
\def\pprw{8.5in}
\def\pprh{11in}

\definecolor{linkColor}{RGB}{6,125,233}
\hypersetup{%
  pdftitle={\plaintitle},
 pdfauthor={\plainauthor},
  pdfkeywords={\plainkeywords},
  pdfdisplaydoctitle=true, 
  bookmarksnumbered,
  pdfstartview={FitH},
  colorlinks,
  citecolor=black,
  filecolor=black,
  linkcolor=black,
  urlcolor=linkColor,
  breaklinks=true,
  hypertexnames=false
}

\begin{document}

\title{\plaintitle}

\numberofauthors{6}
\author{%
  \alignauthor{Natan Liss \thanks{equal contributors}\\
    \affaddr{Technion}\\
    \affaddr{Haifa, Israel}\\
    \email{lissnatan@campus.technion.ac.il}}\\
  \alignauthor{Chaim Baskin \footnotemark[1]\\
    \affaddr{Technion}\\
    \affaddr{Haifa, Israel}\\
    \email{chaimbaskin@cs.technion.ac.il}}\\ 
  \alignauthor{Avi Mendelson\\
    \affaddr{Technion}\\
    \affaddr{Haifa, Israel}\\
    \email{avi.mendelson@cs.technion.ac.il}}\\    
  \alignauthor{Alexander M.Bronstein\\
    \affaddr{Technion}\\
    \affaddr{Haifa, Israel}\\
    \email{bron@cs.technion.ac.il}}\\ 
  \alignauthor{Raja Giryes\\
    \affaddr{Tel-Aviv University}\\
    \affaddr{Tel-Aviv, Israel}\\
    \email{raja@tauex.tau.ac.il}}\\ 
}

\maketitle

\begin{abstract}
Convolutional Neural Networks (CNN) has become more popular choice for various tasks such as computer vision, speech recognition and natural language processing. Thanks to their large computational capability and throughput, GPUs ,which are not power efficient and therefore does not suit low power systems such as mobile devices, are the most common platform for both training and inferencing tasks. Recent studies has shown that FPGAs can provide a good alternative to GPUs as a CNN accelerator, due to their re-configurable nature ,low power and small latency. In order for FPGA-based accelerators outperform GPUs in inference task, both the parameters of the network and the activations must be quantized.
While most works use uniform quantizers for both parameters and activations, it is not always the optimal one, and a non-uniform quantizer need to be considered. In this work we introduce a custom hardware-friendly approach to implement non-uniform quantizers. In addition, we use a single scale integer representation of both parameters and activations, for both training and inference. The combined method yields a hardware efficient non-uniform quantizer, fit for real-time applications. We have tested our method on CIFAR-10 and CIFAR-100 image classification datasets with ResNet-18 and VGG-like architectures, and saw little degradation in accuracy.
\end{abstract}


\section{Introduction}
\label{chap:introduction}
\csname @Latintrue\endcsname 

\subsection{Neural Networks on custom hardware}
When implementing systems involving arbitrary precision,
	FPGAs and ASICs are a natural selection as target device
	due to their customizable nature. It was already shown that there is a lot of redundancy when using floating point representation in Neural Network(NN). Therefore, custom low-precision representation can be used with little impact to the accuracy. Due to the steadily increasing on-chip memory
	size (tens of megabytes) and the integration of high bandwidth memory (hundreds of megabytes), it is feasible to fit
	all the parameters inside an ASIC or FPGA, when using low bitwidth. Besides the obvious advantage of reducing the
	latency, this approach has several advantages: power consumption reduction and smaller resource utilization, which
	in addition to DSP blocks and LUTs, also includes routing
	resource. The motivation of quantizing the activations is
	similar to that of the parameters. Although activations are
	not stored during inference, their quantization can lead to major saving
	in routing resources which in turn can increase the maximal
	operational frequency of the fabric, resulting in increased
	throughput.
		In recent years, FPGAs has become more popular as an inference accelerator. While ASICs \cite{7551407,DBLP:journals/corr/JouppiYPPABBBBB17} usually offers more throughput with lower energy consumption, they don't enjoy the advantage of reconfigurability as FPGAs. This is important since neural network algorithm evolve with time, so should their hardware implementation. Since the implementation of neural network involves complex scheduling and data movement, FPGA-based inference accelerators has been described as heterogeneous system using OpenCL \cite{DBLP:journals/corr/WangAX16,7848692,DBLP:journals/corr/AydonatOCLC17} or as standalone accelerator using HLS compilers \cite{7869873,DBLP:journals/corr/UmurogluFGBLJV16,Zhao:2017:ABC:3020078.3021741}.

\subsection{The effect of quantization choice on hardware}
It is known that FPGAs and ASICs are good with custom
data types arithmetic and logical operations, especially integers and fixed point. During inference, only activations are
actually being quantized since the parameters stored in the
memory have already been quantized during training. The
simplest and probably the most hardware efficient implementation of activation quantization is the linear quantization, which can be easily implemented by applying bit shifting on
the result of Multiply And Accumulate (MAC) operation,
after all the negative values have been zeroed. The size
of the shift depends on the maximum value of the MAC
operation and the number of bits representing the activation.
Suppose the activation is represented as $BW_{a}$ a bit unsigned
integer and the network parameters are represented as $BW_{p}$
bit signed integer. Also, suppose we have a filter with a size
of $N_f$ . The amount of right bit shift required for linear quantization is: $\log_2[(2^{BW_a}-1)(2^{BW_p-1}-1)N_f]-BW_a $.
Another hardware efficient activation is the logarithmic. When
using the $log_2$ , the implementation simply boils down priority encoder on the MAC result which indicates the location
of the most significant '1'. The major drawback of  $log_2$ based activation is that the gradient can diminish rapidly, which can lead to slow convergence, especially for the first layers. A more general form of quantized activation can be formulated using thresholds
example when using non-uniform quantization, which in hard-
ware is realized with chains of comparators and MUXs.

\section{Related Work}
Previous works had been investigating the challenging task of efficiently deploying neural network(NN) to a custom hardware. This task usually boils down to two major sub-tasks:
\begin{itemize}
  \item Network quantization.
  \item Pipelining and scheduling.
\end{itemize}
Extreme quantization setup, as low as 1-2 bit weights and activation quantization, has been explored \cite{Baskin2017StreamingAF, rastegari2016xnor,hubara2016quantized, hubara2016binarized, DBLP:journals/corr/CourbariauxB16, zhou2016dorefa} In some of these works, the multiplying operation has been replaced by a sequence of XNOR and Pop-count operations which increase the custom logic utilization of an FPGA or ASIC. 
However, this works suffer from accuracy drop due to the low precision. In \cite{zhu2016trained} work, the authors address this issue by  adding a scaling layer after a quantized layer as well as learning two independent weight thresholds. while \cite{polino2018model} approach this by  widening the layer depth, at the cost of higher computational effort.
Recently \cite{mishra2017apprentice} and \cite{polino2018model} proposed, a teacher-student setup for knowledge distillation \cite{hinton2015distilling}. In this setup, a full-precision model is used to distill its knowledge to the target network. While this approach achieves impressive results, the training computation complexity increases by more than $2X$.
In the aforementioned papers, the general approach is to preform the forward pass with the quantized weight values, while the backward pass is preformed on a full-precision copy of the weights.
In \cite{DBLP:journals/corr/abs-1712-05877}, the authors represent each parameter as an 8-bit integer coupled with layer-specific scaling and zero-point factors. Using this quantization approach allows integer-only inferencing, with little drop in accuracy.    
A common approach for aforementioned paper, and most previous works on quantization, assume a uniform distribution of weights and activation. While this assumption leads to a more compelling hardware implementation. Unfortunately, the distribution of weights looks like they were drawn from gaussian distribution \cite{han2015deep,Anderson2018High}.
Following this assumption, \cite{DBLP:journals/corr/abs-1804-10969} suggested to first calculate $\mu$ and $\sigma$ per each layer and apply CDF of the normal distribution, which subsequently transform the parameters from normal to uniform distribution. A noise, drawn from uniform distribution, is added to simulate the quantization error. Finally, a quantile function is applied in order to transform back the parameters to normal distribution.
In recent years, FPGAs has become more popular as an inference accelerator. And while ASICs \cite{7551407,DBLP:journals/corr/JouppiYPPABBBBB17} usually offers more throughput with lower energy consumption, they don't enjoy the advantage of reconfigurability as FPGAs. This is important since neural network algorithm evolve with time, so should their hardware implementation. Since the implementation of neural network involves complex scheduling and data movement, FPGA-based inference accelerators has been described as heterogeneous system using OpenCL \cite{DBLP:journals/corr/WangAX16,DBLP:journals/corr/AydonatOCLC17,7848692} or as standalone accelerator using HLS compilers \cite{DBLP:journals/corr/UmurogluFGBLJV16,Zhao:2017:ABC:3020078.3021741, 7869873}

\section{Non uniform quantization hardware implementation}
\subsection{Activation quantization using thresholds}	
In recent years, ReLU has been a popular choice for activation function because it allows faster convergence of deep networks. This activation function also appeals  for custom hardware such as FPGAs \cite{DBLP:journals/corr/UmurogluFGBLJV16,DBLP:journals/corr/WangAX16,losslessly_quantized} and ASICs since the implementation involves 1-bit compactor for the sign bit and a MUX which either passes zero-vector or the input. However, the fact that ReLU is unbounded coupled with the the large MAC results, imposed by integer only networks, makes low bit quantization very challenging. 
We take a reasonable assumption that the activations follows a Gaussian distribution. In order to achieve equiprobable distribution among all bins, we use a hybrid of CDF of a normal distribution and ReLU. Let $T = \{0 < t_0 < t_1 < \cdots < t_{k-3} < t_{k-2} = \infty\}$ be a set of thresholds and $B_a$ the number of allocated bits for quantized activation. We will define a quantizer $Q(x)$ which maps the result of MAC operation $x \in \mathbb{Z}$ to $y \in \mathbb{Z}$
 according to the following :
\[
y = Q(x)=
\begin{cases}
  0, & x \in (-\infty,0]\\
  1, & x \in (0,t_{1}]\\
  \vdots              \\
  2^{B_a}-1, & x \in (t_{k-2},t_{k-1}]\\
  2^{B_a}-1, & x > t_{k-1}
\end{cases}
\]
During train time, we calculate the running statistics, $\mu$ and $\sigma$, for each layer, and use them during evaluation to calculate the thresholds $T$.
Let $F_X(x,\mu,\sigma)$ be the CDF of a normal distribution and $F_X^{-1}(x,\mu,\sigma)$ be the quantile function of a normal distribution.
Activation thresholds $T$ are calculated as follows:
$$
Z=F_X(0,\mu , \sigma )
$$
$$
b_i  = \left[\frac{1-Z}{2^{B_a}} \right] i , \qquad i \in[1\mathrel{{.}\,{.}}\nobreak 2^{B_a}-1]
$$

$$
t_i = F_X^{-1}(b_i, \mu, \sigma) 
$$

\begin{figure}[t]
			\centerline{\includegraphics[width=\linewidth]{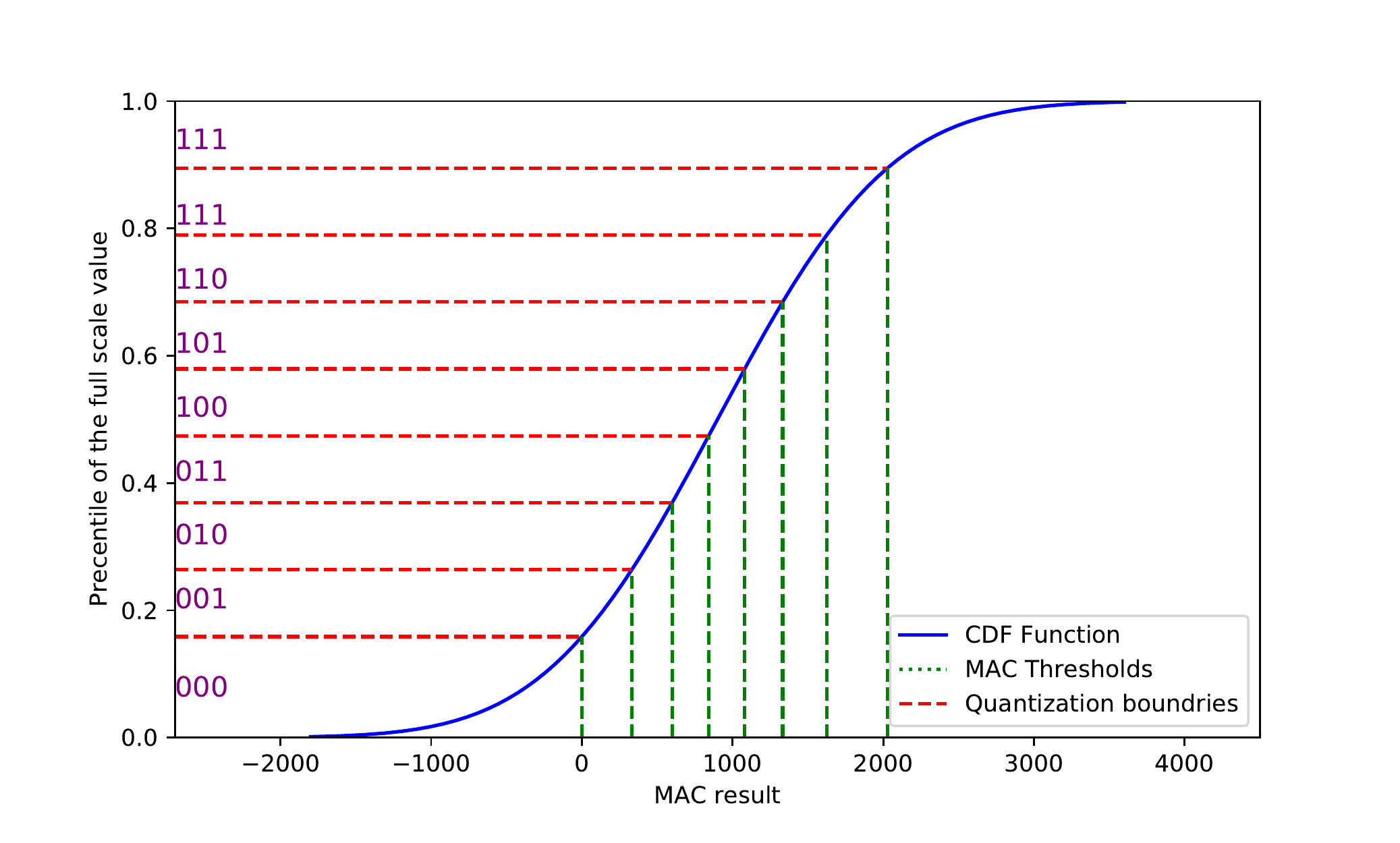}}
			\caption{k-quantile quantization to discrete integer levels}
			\label{cdf_actquant}
	\end{figure}

An illustration of this function, for activation with $\mu=900$ and $\sigma=900$, is present in Figure \ref{cdf_actquant}. From this we can clearly see that the distance on y-axis are equal which implies equiprobable bins. The intersection of the vertical green lines with the x-axis represent the integer thresholds which are loaded to the custom device, during inference, and define the boundaries between each bin.

\subsection{Non uniform integer weight quantization}
Following the assumption that weights are normally distributed,  we used a similar non uniform quantizer as mentioned above.
Let $F_X(x,\mu,\sigma)$ be the CDF of a normal distribution and $B_w$ the number of allocated bits for quantized weights. We quantize weights as follows:
$$
W_q =  \left[F_X(W,\mu_w,\sigma_w)-0.5\right]\times 2^{b_w} 
$$

\subsection{Back propagating through Gaussian threshold activations quantizer}
Since $Q(x)$ is a non differentiable function, we used the Straight-Through
Estimator (STE) \cite{bengio2013estimating} approach in order to approximate the gradient of a quantized variable. Our chosen activation function is the normal CDF $F_X(x,\mu(x),\sigma(x))$. Combining this fact with the STE approach, the combined derivative of the CDF activation and quantization functions w.r.t input $x$, is the input derivative multiplied by normal probability density function: $F'_X(x,\mu(x),\sigma(x))$

\subsection{Residual block adaptation}
\label{residual_block_adaptation}
\begin{figure}
        \begin{subfigure}[t]{0.4565\linewidth}
            \includegraphics[width=\linewidth]{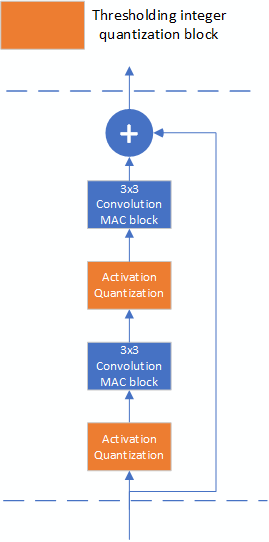}
            \caption{Custom residual basic block for integer quantization} 
            \label{custom_basic_block}
        \end{subfigure}
        \hspace{\fill} 
        \begin{subfigure}[t]{0.5017\linewidth}
            \includegraphics[width=\linewidth]{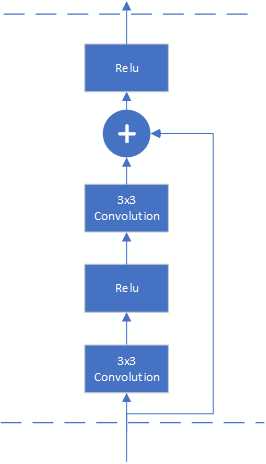}
            \caption{Original residual basic block} 
            \label{original_basic_block}
        \end{subfigure}
        \caption{Custom vs original residual basic block architecture} \label{original_vs_basic_block}
    \end{figure}

In order to be able to incorporate integer-only arithmetic with residual based architectures \cite{He_2016_CVPR}, we had to do slight modification in the residual block.
The modification to the basic block is visible in Figure \ref{original_vs_basic_block}.
In the original basic residual block, the last activation function is applied after adding the result of the convolution layer with the residual result, as shown in Figure  \ref{original_basic_block}. If we simply replace the second ReLU function with integer quantization, this would bound the residual of the next layer to the maximum value of  $M_a = 2^{BW_a}-1$ where $Ba$ is the number of bits allocated for quantized activation. For the second convolution layer, of the  dimensions $O\times I \times K^{2}$ where $O,I,K^{2}$ are the number of output feature maps, input feature maps, and size of filter respectively, the maximal MAC value is $M_m = (2^{BW_a}-1)(2^{BW_a-1}-1)(I\times K^{2})$.
Since $M_m >> M_a$, the effect of the residual path will become negligible and eventually lead to accuracy drop.
To tackle this issue, we have moved the post-residual activation to the input of the next basic block, as shown in Figure \ref{custom_basic_block}. In this case, the scale of the residual and the MAC result of the second convolution layer is roughly the same.


	
\section{Experimental results}
    \subsection{Training setup}
    In order to avoid model divergence, which may occur when all of the layers' parameters are quantized, we follow a gradual quantization scheme, similar to the one proposed by \cite{AAAI1816479}.
    The gradual training works as follows: Lets denote the total number of layers as $N$. At the $i$-{th} stage, the parameters and activations of layers $\{L_1,...,L_i\}$ are quantized, while layers $\{L_{i+1},...,L_N\}$ remain in full precision. This training scheme allows smoother and more stable convergence to a point in which all the network is quantized. As a representative of non-uniform quantization function, for both parameters and activations, we chose the normal CDF function. 
    We use an SGD optimizer with momentum. The learning rate is set to $10^{-3}$, momentum $0.9$ and weight decay $10^{-4}$.

    \subsection{Results}
    We have tested threshold quantization on CIFAR-10 and CIFAR-100 image classification datasets using ResNet-18 and VGG-like architectures. The residual blocks in the ResNet-18 architecture, were modified as described in previous sections. Our VGG-like architecture is shown in Table \ref{vgg_like_arch}. The results for CIFAR-10 and CIFAR-100 datasets, are presented in Table \ref{cifar_thresholding} and Table \ref{cifar100_thresholding} accordingly.
    The rows with the (w,a)=32,32 setup, represent the baseline accuracy. As one can observe, there is a little degradation in accuracy, for both datasets. We did observe slightly bigger degradation of VGG-like architecture for CIFAR-100 dataset, which may be attributed to the overall weakness of these types of architecture in term of diminishing gradients and ability to adapt to quantization via gradient descent. The ResNet-18, thanks to their skip connection, tend to fine tune batter to quantization.

    \begin{table}[H]
    \centering
    	\caption{VGG-like architecture for CIFAR-10. Convolution layers are denoted as "conv\big \langle receptive field size\big \rangle-\big \langle number of channels\big \rangle"}
        \label{vgg_like_arch}
        \begin{tabular}[c]{|c|}
            \hline
            conv3-128\\ \hline
            conv3-128\\ \hline
            maxpool  \\ \hline
            conv3-256\\ \hline
            conv3-256\\ \hline
            maxpool  \\ \hline
            conv3-512\\ \hline
            conv3-512\\ \hline
            maxpool  \\ \hline
            FC-1024  \\ \hline 
            FC-10  \\ \hline 
        \end{tabular}
    \end{table}

    \begin{table}[H]

		\centering
		\caption{Threshold based integer-only quantization results for CIFAR-10 dataset}
		\begin{tabular}{ccc}
			\toprule
			\textbf{Network} & \textbf{Precision (w,a)} & \textbf{Accuracy (\% top-1)} \\ \midrule
			ResNet-18 &    32,32    & 93.76     \\ 
			ResNet-18 &    4,4      & 93.45     \\
			ResNet-18 &    3,3      & 93.05     \\ \midrule
			VGG-like &     32,32    & 90.5     \\ 
			VGG-like &     4,4      & 89.7     \\
			VGG-like &     3,3      & 88.1     \\
			\bottomrule
		\end{tabular}
		
		\label{cifar_thresholding}
	\end{table}
	
	\begin{table}[H]
		\centering
		\caption{Threshold based integer-only quantization results for CIFAR-100 dataset}
		\begin{tabular}{ccc}
			\toprule
			\textbf{Network} & \textbf{Precision (w,a)} & \textbf{Accuracy (\% top-1)} \\ \midrule
			ResNet-18 &    32,32    & 73.95     \\ 
			ResNet-18 &    4,4      & 71.71     \\
			ResNet-18 &    3,3      & 70.50     \\ \midrule
			VGG-like &     32,32    & 67.1     \\ 
			VGG-like &     4,4      & 64.5     \\
			VGG-like &     3,3      & 63.1     \\
			\bottomrule
		\end{tabular}
		\label{cifar100_thresholding}
	\end{table}
	
\section{Conclusion and future work}
In this work we have presented a hardware efficient technique for implementing non-uniform quantizer. From Tables \ref{cifar_thresholding} and \ref{cifar100_thresholding} it is clear that using integer parameters and activations leads to almost no degradation in accuracy, for CIFAR-10 and CIFAR-100 datasets, and can be efficiently implemented costume hardware. In the future we would like to implement threshold based non-uniform quantizer on an FPGA and compare it to straight forward implementation, from logic utilization, power and runtime perspective.    

\balance{}

\bibliographystyle{SIGCHI-Reference-Format}
\bibliography{sample}

\end{document}